\documentclass[runningheads]{llncs}

\usepackage{graphicx}
\usepackage{hyperref}

\begin{document}

\title{Towards Knowledge Organization Ecosystems}
\titlerunning{Towards Knowledge Organization Ecosystems (KOEs)}

\author{Mayukh Bagchi}

\authorrunning{Mayukh Bagchi}

\institute{Department of Information Engineering and Computer Science (DISI), \\ University of Trento, Trento, Italy \\ \email{mayukh.bagchi@unitn.it}}

\maketitle              

\begin{abstract}
It is needless to mention the (already established) overarching importance of knowledge organization and its \emph{tried-and-tested} high quality schemes in knowledge based Artificial Intelligence (AI) systems. But equally, it is also hard to ignore that, increasingly, standalone KOSs are becoming \emph{functionally ineffective} components for such systems, given their \emph{inability} to capture the continuous facetization and drift of domains. The paper proposes a radical re-conceptualization of KOSs as \emph{a first step to solve such an inability}, and, accordingly, contributes in the form of the following dimensions: (i) an explicit characterization of Knowledge Organization Ecosystems (KOEs) (possibly for the first time) and their positioning as pivotal components in realizing sustainable knowledge-based AI solutions, (ii) as a consequence of such a novel characterization, a first examination and characterization of KOEs as Socio-Technical Systems (STSs), thus opening up an entirely new stream of research in knowledge-based AI, and (iii) motivating KOEs not to be mere STSs but STSs which are grounded in Ethics and Responsible Artificial Intelligence cardinals from their very genesis. The paper grounds the above contributions in relevant research literature in a distributed fashion throughout the paper, and finally concludes by outlining the future research possibilities.

\keywords{Knowledge Organization Ecosystems \and KOEs \and Knowledge Representation \and Socio-Technical Systems \and AI Ethics \and KOEs as Ethics-Aware STSs}
\end{abstract}

\section{Introduction}
\label{sec:Introduction}
Over many decades, organizing and representing knowledge resources and providing access to them through a subtle combination of knowledge organization tools has always been a passion and a challenge for actors involved in different information-central professions. Knowledge Organization Systems (KOSs), as of today, mostly work in a \emph{make-do} fashion. For example, for any knowledge domain, term lists (such as authority files, glossaries) establishes standardized terminology in that domain with their definitions, classification and categorization schemes imposes a hierarchical (broadly, taxonomic) backbone to their key concepts, relationship lists (such as thesauri, classification ontologies) facilitates the intra and inter semantic linkages between those concepts, and cataloging codes and metadata standards makes such knowledge discoverable \cite{Zeng2008,Birger2008}. But, in \emph{``the present era of big data and information explosion, domains are becoming increasingly intricate and facetized, often leaving traditional approaches of knowledge organization functionally inefficient in dynamically depicting intellectual landscapes'} \cite{Bagchi2021}. Such incoherence is especially pronounced in the context of knowledge-based AI systems due to three prime reasons. Firstly, KOSs, as of now, are (mostly) studied as standalone knowledge artifacts instead of being examined as an expanded ecosystem where individuals and organizations bear as much, if not more, importance for sustainable knowledge modelling and management. Secondly, as a consequence of the first reason, there is no (even preliminary) explication of the potential role of individuals and organizations in sustainable knowledge modeling. Lastly, there is (almost) a total absence of ethical considerations in designing KOSs (historically but also as of today) which is of particular concern as KOSs are playing an increasingly pivotal role in knowledge-based AI and data intelligence. To that end, the paper considers a radical re-conceptualization of KOSs by proposing the following \emph{novel} contributions:-

\begin{enumerate}
    \item a paradigm shift from Knowledge Organization Systems (KOSs) to Knowledge Organization Ecosystems (KOEs), the shift being decisive for achieving \emph{sustainable} knowledge-based artificial intelligence,
    \item leveraging the former, a first examination and characterization of Knowledge Organization Ecosystems (KOEs) as Socio-Technical Systems (STSs), thus opening up a new dimension of research in knowledge-based AI
    \item finally, to ground the fact that KOEs are not mere STSs but STSs which are designed following AI ethics cardinals from their very inception.
\end{enumerate}

\noindent In the context of the above contributions, let us see how the graded characterization of KOEs are pivotal for designing knowledge-based AI systems as components of larger AI ecosystems. Firstly, knowledge-based AI systems (ideally) require knowledge models which are modular and evolving (in short, sustainable) in nature, and the shift from KOSs to KOEs facilitates just that by linking standalone knowledge models to individuals and organizations which drive their evolution. But stopping at such a characterization isn't enough as it doesn't explicate the specific components of each- individuals, technology and organizations, which cumulatively comprise the expanded and continuously evolving knowledge design required in today's AI systems. Such a characterization is achieved by envisioning KOEs as Socio-Technical Systems (STSs). Finally, the KOEs are characterized as different from mainstream STSs as they are normatively guided by AI ethics cardinals, which, though long overdue, are recently getting research traction (but hardly amongst knowledge organization and representation-based AI systems and architectures).

The remainder of the paper is organized as follows: Section 2 elucidates the paradigm shift from Knowledge Organization Systems (KOSs) to Knowledge Organization Ecosystems (KOEs). Section 3 focuses on a first characterization and examination of KOEs as Socio-Technical Systems (STSs). Section 4 emphasizes on the importance of designing ethical KOEs from their very inception, and towards that briefly introduces and contextualizes three AI ethics cardinals- \emph{cognitive biases, metadata for explainability} and \emph{datasheets for transperancy, accountability and refinement}. Finally, Section 5 summarizes the entire work and particularizes the future research directions.

\section{From KOSs to KOEs}
\label{sec:From KOSs to KOEs}
Knowledge Organization Systems (KOSs), \emph{``schemes for organizing information and promoting knowledge management"} \cite{Zeng2008}, have been the object of perennial research inquiry within the premises of Knowledge Organization (KO) \cite{Birger2008} and Knowledge Representation (KR) \cite{Bagchi2021}, facilitating \emph{``resource discovery and retrieval by acting as semantic road maps...for indexers and future users, either human or machine"} \cite{Zeng2008}. The spectrum of established genres of KOSs are quite diverse with respect to communities of practice - generic, interdisciplinary or domain-specific knowledge models - as well as in their structuring, ranging from flat structures like glossaries and synonym rings to multidimensional knowledge models such as semantic networks, ontologies, and now, Knowledge Graphs (KGs) \cite{Bagchi2019b,Hogan2020}. Further, from the utility perspective, KOSs have been employed by librarians for organizing library resources, by solution architects for data integration and interoperability in semantic knowledge management architectures \cite{kgcw}, by computational linguists for Natural Language Processing (NLP) \cite{Miller1995}, and as back-end semantic technologies in facilitating conversational intelligence \cite{bagchi2020b}.

A careful examination of such KOSs\footnote{\url{https://www.isko.org/cyclo/kos}}\footnote{\url{https://nkos.slis.kent.edu/2001/SoergelCharacteristicsOfKOS.pdf}} ascribes several generic (ideal) characteristic features to such schemes. \emph{Firstly}, the \emph{cognitive grounding} of KOSs as uniformly inferred from the various theories of mental representation\footnote{\url{https://plato.stanford.edu/entries/mental-representation/}}, and the \emph{continuous inductive nature} of such grounding being exceedingly important in light of: (1) continuous facetization and growth of the universe of subjects, knowledge and domains \cite{srr1967}, and (2) the implicit requirements put forth by paradigms in \emph{evolutionary computation} \cite{Thomas1997}. \emph{Secondly}, the commitment (such as in \emph{ontological commitment} \cite{Pisanelli2004}) embodied in the \emph{conceptual infrastructure} of the KOSs is crucial for the practical employment of such a system, meaning by this a pragmatic coverage of the representational needs fulfilling the final purpose of the artifact as in, for instance, intelligent information systems (especially important in precision and niche domains like aviation information systems, wherein, often there is a need to etch a fine-grained knowledge model on, for example, \emph{`Flight Plans'}, and in doing so, to keep its coverage \emph{representationally distinct} yet \emph{functionally connected} to other representational sub-systems in an aviation ecosystem such as Weather Forecast, Cargo Handling, Flight Maintenance, Catering etc). \emph{Thirdly}, and complementary to the first and second characteristic, is how \emph{realistic} such a scheme of KO is with respect to the ground knowledge as evident in different requirement scenarios such as intellectual domain modelling and data intelligence (for example, how realistic a KOS is in structuring knowledge of ancient Indian socio-religious order when designing it for realizing a digital library on Vedanta Philosophy). {Finally}, and in consonance with all of the above characteristics, the \emph{implementational flexibility} of such a KOS as examined from the prism of information storage, retrieval and technology stacks.

The characteristics of KOSs which were adumbrated above raises several \emph{non-trivial} issues, especially when considering them from the overarching perspective of Networked KOSs\footnote{\url{https://nkos.slis.kent.edu/2019NKOSworkshop/JDIS-SI-NKOS-Full.pdf}} and Linked Data \cite{Bizer2011} in knowledge organization and representation backed intelligent systems. The first issue relates to the family of \emph{cognitive biases}\footnote{\url{https://plato.stanford.edu/entries/implicit-bias/}} (such as affinity bias, implicit bias) which are (possibly) inherited by KOSs via their inherent grounding in human \emph{cognition} and \emph{consciousness}. It is of utmost importance to keep in check the percolation of such \emph{pre-emptive ideation} about any aspect of a domain of discourse, especially in consonance with the emerging research vibrancy around ethical and responsible AI\footnote{\url{http://aiethics.site/}} (for instance, an interesting case would be a generic KOS of data science which is completely oblivious, or in other words, biased against its interdisciplinary kernels, such as from Library and Information Science, Sociology etc). The same example can be utilized to focus on another issue, that of \emph{user feedback}, wherein, even if the user(s) sense any bias or disequilibrium in commitment in the KOS for the purpose they are using, there is no systematic consideration of revision and request for committing the revision in a time-bound, peer-reviewed fashion. Another two issues which are closely interrelated as well as related to user feedback, are those of \emph{semantic drift}, which in the parlance of KO means \emph{``that a concept is gradually understood in a different way or its relationships with other concepts are undergoing some changes"} \cite{Gulla2010}, and the non-consideration of such a drift resulting in minimal to almost no \emph{reuse} of such KOSs. Finally, the need for an organizational anchor to ground, manage and accommodate all these issues resulting in \emph{multidirectional inheritance}.

What we notice from the above threads of discussion is the need for an \emph{explicit mutualism} amongst people, technology and organizations (and the numerous sub-systems therein) with respect to knowledge modelling for knowledge-based AI systems, and an \emph{obfuscation} regarding the characterization of KOSs in such terms as of today. The advocacy is for a \emph{paradigm shift from the singular study of Knowledge Organization Systems (KOSs) to Knowledge Organization Ecoystems (KOEs) - studying KOSs as a larger triad of people, technology and organizations}. Accordingly, Knowledge Organization Ecoystems (KOEs) are conceived as an ecosystem of interdependent components - individuals, knowledge organization models and organizations - that collaborate and combine their individual efforts towards a more \emph{coherent} and \emph{sustainable} knowledge management solution (as distinct from pure knowledge management models \cite{McAdam2005} where KO doesn't necessarily play a role). Such a characterization of KOEs can lead to several ramifications, of which few prominent ones are highlighted as follows-

\begin{enumerate}
\item The \emph{first} being the general \emph{evolutionary development} \cite{JMG2018} of such an ecosystem arising out of the \emph{symbiosis} amongst various social, organizational and technological components (accommodating the changing nature of cognitive conceptualization). 
\item Following the very existence of such KOEs, the establishment of a consensual body of canons for their \emph{governance}, a section of which should ideally strive to identify and strategically mitigate issues such as cognitive biases. 
\item The KOEs will, as a result of such characterization, also manifest the dual qualities of \emph{modularity} - which will allow each component and sub-component in the ecosystem to be autonomous in their own design, and \emph{complementarity} - the autonomous modules to be functionally dependent on each other. 
\item \emph{Value Capture} is another fundamental kernel of KOEs - with the essence being that the knowledge captured in KOEs are to be treated as open assets for reuse and remixing (the ecosystem being grounded in open science \cite{os2014}). 
\item Last but not the least, KOEs will usher in true \emph{multi-dimensionality in knowledge modeling}, in the sense that a single KOE (for instance, on aviation) can comprise of several genres of knowledge models with varying dimensionality (like, depth classification schemes for different components of aviation; ontologies, and subsequently KGs designed by utilizing such robust classification schemes as their base etc.)
\end{enumerate}
The KOEs are thus envisioned to possess living, biological ecosystem like attributes which will make them truly \emph{``a set of actors with varying degrees of multilateral, nongeneric complementarities that are not fully hierarchically controlled"}\footnote{A detailed parametric characterization of KOEs grounded in the theory of ecosystems will be made in a future paper, with this being a first preliminary sketch.} \cite{JMG2018}.

\section{Characterizing KOEs as STSs}
\label{sec:Characterizing KOEs as STSs}
The paradigm shift from KOSs to Knowledge Organization Ecosystems (KOEs) accredits the (further) characterization of KOEs as Socio-Technical Systems (STSs), which, quoting from \cite{salnitri2014}, are \emph{``complex systems where social (human and organizational) and technical components interact with each other to achieve common objectives"}. It is clear from the prefaced definition (and also inferred from similar attestations in \cite{Cooper1971,Maguire2014}) that any STS rests upon three foundational cardinals - Individuals, Organizations, and Technology Components - which are also the three fundamental pillars instituting any KOE. The proposal in this section is to establish a conceptual mapping between STS theory and KO theory with respect to these three key components which are central to both, and thus componentially induce a first characterization of KOEs as STSs in the overall context of knowledge-based AI. An interesting observation of such a characterization would be to pinpoint the sub-components within the three pillars which are unique to KO and which are usually not considered essential for mainstream STSs.

\subsection{Individuals} 
Individuals constitute the most important pillar of any KOE given that it is ultimately they who design them for for their own varied purposes and applications. To that end, there are five mutually inexclusive categories of individuals who should ideally play a collaborative role in sustaining KOEs - Designers, Domain Experts, Transdisciplinary Ethics Experts, Users and Datasheet Documentalists. 
\begin{enumerate}
    \item \emph{Designers} are central to KOEs because they are theoretically and technically capable to understand and implement the nuances of actually devising knowledge artefacts in consonance with domain and ethics experts. To take an example, the \emph{core task} of designing domain ontologies and knowledge graphs can \emph{only} be delegated to ontology engineers since such a design requires considerable theoretical foundations (such as in taxonomy, mereology, causality, philosophy of language, mathematical logic) coupled with technical know-how (such as modelling languages, schema design tools, ontology modelling tools, graph databases). The confluence of such theoretical and technical knowledge also makes designers perfectly qualified to implementationally etch crucial design choices such as ontological commitments.
    \item \emph{Domain Experts}, on the other hand, contribute in providing crucial input regarding the conceptual modelling of the domain, application area, task or process, of which semi-formal or formal semantics is being modelled and codified. For instance, it would be near to impossible for a classificationist and an ontology engineer to design a classification scheme and subsequently a formal classification ontology for Indian Hindu festivals without inputs from a domain expert who can be an established academician or researcher, or even be highly regarded professionally in that domain. The only exception to this convention can be when the designer couples as a domain expert (which is possible but observationally infrequent). 
    \item \emph{Transdisciplinary Ethics Experts} are of overarching importance in this setup because they are adept at locating the ethical issues, concerns and questions which are essential to embed fairness, transparency and explainability in the knowledge models being developed. It is important to mention that such expertise in ethics should be transdisciplinary in nature, making them fit for collaborating with groups in, for instance, different domains. To take an example, the issue of diverse genres of affinity bias such as partisanship towards western philosophy backed knowledge structuring (especially in religious and sexual orientation; see \cite{bullard2020} for the latter) in established classification schemes are crucial ethical hotspots where ethics experts should facilitate remediation (more so, as classification schemes are often taken as base models for developing ontologies and KGs). \item In addition to these three categories, the fourth category of individuals, the \emph{Users} constitute the most important category in the sense that it is ultimately they who will practically utilize the artifacts for the application goals towards which they are working. In this context, it is important to note that a systematic process of incorporating \emph{user feedback} is vital towards successful versioning, and ultimately sustenance of the knowledge models being developed. 
    \item Finally, \emph{Datasheet Documentalists} will be responsible for continual generation of datasheets \cite{gebru} for the KOEs documenting their motivation, composition, uses, distribution and maintenance by interacting with responsible individuals from other categories, or to say alternatively, be responsible for the centralized knowledge management of KOEs. It is important to stress that the role of datasheet documentalists are pivotal for pragmatically effectuating KOEs which conform to AI ethics cardinals from their very inception.
\end{enumerate}

\subsection{Technology} 
Technology is central to KOEs and accordingly, there are two simultaneous axes to the technology dimension of KOEs as STSs- \emph{Vertical Axis} and \emph{Horizontal Axis}. The \emph{Vertical Axis} deals with the knowledge organization and representation artifacts which are domain-specific, application-specific or task-specific. For this axis, the overarching focus is on the ontological commitment embodied in such knowledge models as the representational appropriateness of such commitment will determine the impelementational worthiness of such models. To that end, the first set of tools which are pivotal for developing a (semi-formal) blueprint of the intended knowledge model are the architecture design tools (open source diagramming tools). The second crucial requirement, more relevant for multidimensional knowledge models, remains the essentiality of a modelling language and subsequently tools which aids designers to utilize such languages to formally encode the model architecture. Thirdly, though an optional requirement, the increasing relevancy of employing and customizing information visualization tools to visualize the landscapes in knowledge models (see \cite{Bagchi2019a} for the importance of domain visualization). 

The \emph{Horizontal Axis}, on the other hand, focuses on the inter and intra domain knowledge artifacts and how they inform and are functionally dependent on each other, especially as seen in the light of Linked Data interoperability requirements \cite{Bizer2011}. To take an example, top-level ontologies are often employed in the ontology engineering community to functionally unify domain or application-specific ontologies towards developing knowledge graphs for varied purposes. Datasets, which are often considered pivotal for developing knowledge models (like KGs), are also considered within the horizontal technology axis. Further, the axis also operationalizes the overall version management based technology development platform (like Git) which is absolutely essential for committing, revising and reusing knowledge models in a collaborative fashion amongst different genre of individuals anchored by the organizational consortia. It is important to mention that the technology dimension is highly dependent on individuals for design, ethical and intellectual sustainability and on organizations to ground and manage the overall infrastructure of KOEs.

\subsection{Organizations} 
The organizational anchor envisioned for KOEs is modelled on the notion of technology consortiums (a prime example being that of World Wide Web Consortia or W3C\footnote{\url{https://www.w3.org/}} in short). For KOEs, the proposed approach is to constitute domain-focused KOE consortiums comprised of \emph{Special Interest Groups (SIGs)} such as Design SIG, Ethics SIG etc. and an overall Working Group on Governance, all of them working in consonance with each other on the KOE, which can, as mentioned earlier, preferrably be hosted on an open version management-based software development platform to be administered by the working group on governance. Further, an open \emph{Community Group} of users is also integral to such an organizational anchor by providing vital user feedback by raising issues on such a platform, thus facilitating continuous designing, testing, deploying and revisioning of the knowledge models. Such a dynamic-platform oriented approach facilitates the incorporation of the novel \emph{Continuous Integration / Continuous Delivery (CI/CD)} paradigm in software engineering \cite{cicd2017} within the fold of KOEs. Continuous practices such as those mentioned above are considered implementational benchmarks in software engineering industry and research, and thus ascribes to KOEs the novel quality of frequent and reliable release of new versions of knowledge models in a cyclic fashion. Further, such domain-focused KOE consortiums can then form network of consortiums, thus, in time, realizing the true potentialities (and also issues) arising in an ecosystem framework. \\

\noindent There are four observations which need to be highlighted in view of the above characterization of KOEs as STSs. Firstly, the individual components of individuals, technology and organizations, and many of their sub-components, are not completely novel with respect to KOSs if seen from a high-level view, but their symbiotic characterization in terms of KOEs grounded in socio-technical systems theory, and further, the explicit mutualism of the three components and their subcomponents are definitely novel to mainstream KO (more so, as KOSs are usually studied from a standalone perspective in mainstream KO). Secondly, the innate importance of ethical design norms in the constitution of KOEs is another dimension which is attached very less prominence in mainstream KO but should ideally be part of the foundational design stack in any knowledge-based AI system. It should also be noted as a corollary that most KO and domain development design frameworks often club the roles of designer, domain expert and ethics expert which can be a \emph{rare possibility} but not a \emph{surety}. Thirdly, the introduction of version management based knowledge development platforms for hosting and managing KOEs is another novel feature of the above characterization. Finally, as a consequence of all of the above, the novel characterization of KOEs as a knowledge based CI/CD paradigm.

\section{Grounding KOEs as AI Ethics-aware STSs}
\label{sec:Grounding KOEs as AI Ethics-aware STSs}
For knowledge organization ecosystems to be characterized as intelligent socio-technical systems \cite{jones2013} supporting knowledge-based AI applications, they require an innate localization and consideration of AI ethics cardinals from their very (design) conceptualization, where AI ethics are \emph{``focused on ``concerns" of various sorts, which is a typical response to new technologies"}\footnote{\url{https://plato.stanford.edu/entries/ethics-ai/}}. In this context, it is important to note the fact that we're considering intelligence (as in intelligent systems) as not merely the abilities of logical contextualization, learning and reasoning but also \emph{from the prism of social intelligence} embodied in knowledge-based AI systems, and that is exactly where ethical concerns play an overarching role. The wealth of research literature on responsible and ethical AI is varied, relevant and overwhelming, but for the purpose of KOEs and their possible applications, let us focus on three innovative ethical concerns tailored towards facilitating a second level characterization of KOEs as AI ethics-aware STSs. 

\subsection{Cognitive Bias}
At the outset, let us first focus on the family of cognitive biases, wherein \emph{``human cognition reliably produces representations that are systematically distorted compared to some aspect of objective reality"} \cite{cbias2015}. The particularization and assessment of KOEs with regard to cognitive biases follows the premise that \emph{``if a cognitive bias positively impacted fitness, it is not a design flaw—it is a design feature"} \cite{cbias2015}, or in other words, whether a cognitive bias needs to be mitigated is majorly dependent on the ontological commitment that a KOE or its modules embody. Towards that let us briefly highlight three prominent genres of cognitive biases. \emph{Implicit bias} results in an unconscious, pre-conceived perceptual ideation (positive or negative) about a domain, application or process subsequently instantiated in knowledge models. For instance, the modelling of socio-political structure of any western nation will be flawed if the same is employed to model the socio-political knowledge structure of the state of Manipur in India as it is one of the rare matriarchal societies (thus, reinforcing the collaborative relevancy of domain and ethics experts). \emph{Affinity bias} involves favouring a certain representation with which one is highly familiar, and is observed prominently, for example, while representing knowledge about gender identities, race or religious communities. Such a bias might be a design feature for one-shot KOEs but are glaring flaws if their broad adoption is envisioned. Finally, \emph{Belief bias} implies adherence to a conceptualization that aligns with one's apriori beliefs in opposition to logical thinking and rational understanding. It can often lead to unforeseen consequences, such as the possibility of incorporation of conspiracy theories in knowledge graphs modeling electoral information.

\subsection{Metadata for Explainability}
Next, let us discuss the issue of \emph{metadata management} \cite{bagchi2020a} as the formal foundation for \emph{explainable KOEs} which opens up for consideration a novel ethical cardinal in sync with explainable AI (XAI) \cite{xai}. Explainability is crucial to modern AI systems, including knowledge-based ones, as they \emph{``aim to make AI system results more understandable to humans"} \cite{xai}, and accordingly, a novel (\emph{non-trivial}) proposal is to consider metadata as the pivot which facilitates \emph{machine explainability} in KOEs. There are two highlights legitimizing the argumentation of such a proposal. Firstly, an enriched metadata annotation of KOEs will enable applications (such as predictive analytics applications) subsequently utilizing them to move away from being mere \emph{black boxes} (in which there is no explainability of outcomes) to intermediate \emph{grey boxes}, or even \emph{white boxes} (fully explainable models or applications). Further, in syc with the above argument, the need to justify the outcomes of the applications, or in other words, the principle of \emph{explain to justify} \cite{xai}. To ground its importance, let us assume the scenario of a standalone KG on \emph{aviation catering} which is developed on a specific depth classification scheme designed for only European dietary preferences, and is used as the back-end model for an all-purpose meal recommender system for flyers. It is needless to mention that it'll recommend a very homogeneous meal preference irrespective of the flyer's culinary choices (which, being general purpose, can be as varied as Indian, Latin American). Without a minimum level of metadata annotation (especially, \emph{provenance}) of the KG module, it would be next to impossible to formally decipher why the application is making such a uniform, biased recommendation. The generic composition of the KOEs is ideal to resolve such scenarios as they are (i) are metadata-enriched from design-time throughout their evolution via their inherent version-maanagement based CI / CD architecture, and (ii) the fact that they are ethically founded and evolved.

\subsection{Datasheets for Transparency, Accountability and Refinement}
Finally, let us focus on the ethical importance of documenting KOEs in general, and of tailor-utilizing the novel tool of \emph{datasheets} \cite{gebru} in particular. Our motivation, in sync with their original purpose, is for every KOE to be associated to a datasheet \emph{``that documents its motivation, composition ... recommended uses, and so on"} \cite{gebru}. Differently from metadata management which conveys in a formally encoded fashion the explainability foci of the KOEs, datasheets are of overarching importance for \emph{human explainability} facilitating end-to-end \emph{transparency}, \emph{accountability} and \emph{refinement} of KOEs with respect to their community of practice. To that end, let us briefly discuss the following five mandatory dimensions that datasheets should document for any KOE and their ethical ramifications:
\begin{enumerate}
    \item \emph{Motivation}, wherein the documentalists should primarily articulate, on one hand, the specific reasons behind starting the KOE project (such as motivation, research gaps), whereas, on the other hand, they should clearly articulate the source of the funding as well as competing interests amongst the individuals involved (if any). This dimension anchors transparency and to an extent accountability of the KOE.
    \item \emph{Composition}, detailing the design choices as well as the technical pipelines facilitating the development of the different genres of knowledge models developed within the KOE platform. It is pivotal for future accountability.
    \item \emph{Use}, wherein three complementary aspects should be documented- that of the uses of the KOEs as originally envisaged, the actual usage of the KOEs by the users, and finally, how the above two uses redefine each other in practice. It will be essential for providing valuable inputs regarding the continual refinement of different aspects of the KOE.
    \item \emph{Distribution}, precisely documenting the lifecycle flow, \emph{viz.} pre-alpha, alpha, beta, candidate and stable releases of the KOE during its various phases of development. Thus, it addresses transparency, accountability and refinement cumulatively.
    \item \emph{Maintenance}, elucidating the highlights of design-level, technical, ethical and managerial governance that goes into incremental maintenance of the KOEs. Again, maintenance addresses for KOEs all the ethical cardinals that datasheets ensures.
\end{enumerate}

\section{Conclusion and Future Work}
\label{sec:Conclusion}
The work, in alignment with its stated objectives, established three new dimensions in knowledge organization and representation, crucial for the development of future knowledge-based AI systems. The first part of the paper briefly examined Knowledge Organization Systems (KOSs) as they are studied and used now, and proposed a paradigm shift from the study of Knowledge Organization Systems (KOSs) to Knowledge Organization Ecosystems (KOEs). The second part of the paper leverages the paradigm shift of the first part, and utilizes it towards a first (\emph{brief}) conceptual characterization and component-specific examination of Knowledge Organization Ecosystems (KOEs) as Socio-Technical Systems (STSs). Lastly, the paper argued for characterizing KOEs not as mere STSs but STSs which are grounded in AI ethics cardinals from their very inception. Future work in this arena can revolve around:

\begin{enumerate}
    \item a detailed characterization of KOEs in the interdisciplinary confluence of KO, KR, Knowledge Management (KM) and theory of ecology as from strategic management,
    \item a fine-grained analysis of the correspondence between KOEs and STSs, and in what ways KOEs are different from mainstream STSs,
    \item designing a conceptual framework for KOEs as AI Ethics-aware STSs in the context of knowledge-based AI.
    \item designing KOEs in different domains to examine, validate and extend their impact on knowledge-based AI systems.
\end{enumerate}

\section*{Acknowledgements}
The above research work is an outcome of an independent research project.

\bibliographystyle{splncs04}

\end{document}